\documentclass[runningheads]{llncs}
\usepackage[T1]{fontenc}
\usepackage{graphicx}
\usepackage{algorithmic}
\usepackage{algorithm}
\usepackage{color}
\usepackage{amsmath}
\usepackage{graphicx}
\usepackage{caption}
\usepackage{amsmath}
\usepackage{subfigure}

\usepackage{marvosym}

\begin{document}
\title{Towards High-Performance Exploratory Data Analysis (EDA) Via Stable Equilibrium Point}
%
%
\author{Yuxuan Song* \and
Yongyu Wang*\textsuperscript{\Letter}}
\institute{JD Logistics\\
\email{wangyongyu1@jd.com}}
\maketitle              

\renewcommand{\thefootnote}{}
\footnotetext{*These authors contributed equally and are co-first authors: Yuxuan Song and Yongyu Wang}

\renewcommand{\thefootnote}{}
\footnotetext{\textsuperscript{\Letter}Corresponding author: Yongyu Wang; Correspondence to: wangyongyu1@jd.com}

\begin{abstract}

Exploratory data analysis (EDA) is a vital procedure for data science projects. In this work, we introduce a stable equilibrium point (SEP) - based framework for improving the efficiency and solution quality of EDA. By exploiting the SEPs to be the representative points, our approach aims to generate high-quality clustering and data visualization for large-scale data sets. A very unique property of the proposed method is that the SEPs will directly encode the clustering properties of data sets. Compared with prior state-of-the-art clustering and data visualization methods, the proposed methods allow substantially improving computing efficiency and solution quality for large-scale data analysis tasks.

\end{abstract}
\section{Introduction}
\label{sec:intro}

Exploratory data analysis (EDA) is the procedure of analyzing data sets to gain insights of their characteristics. It is usually the first step for big data analysis. Clustering and data visualization are two main pillars of EDA.

Clustering aims to assign similar data samples into same cluster while assigning dissimilar data samples into different clusters. Spectral clustering is one of the most popular clustering methods. It has superior capability in detecting non-convex and linearly non-separable patterns. However, the computational cost of spectral clustering is very high. Its eigen-decomposition step has a $O(N^3)$ time complexity, where $N$ is the number of data points. So its applicability to truly large scale data analysis is very limited. To solve this problem, substantial effort has been devoted and numerous methods have been proposed: \cite{yan2009fast} proposed to reduce the problem size of eigen-decomposition by performing a coarse-level clustering with $k$-means. \cite{fowlkes2004spectral} attempted to find approximate solution via sampling. Inspired by sparse coding, \cite{chen2011large} proposed to perform spectral embedding via landmark points. However, the existing methods either do not scale, or result in a significant degradation of clustering accuracy.

For the data visualization tasks, t-distributed stochastic neighbor embedding (t-SNE) \cite{van2008visualizing} is the most widely used tool for visualizing high-dimensional data. It aims to learn an embedding from the high-dimensional space to two or three dimensional space in such a way that similar samples are modeled by nearby points in low-dimensional space and dissimilar samples are modeled by distant points so that people can directly view the structure of data sets. However, its involved Stochastic Neighbor Embedding (SNE) process requires to perform gradient descent to minimize the Kullback-Leibler (KL) cost function for every sample which is a very time-consuming process and impractical for large-scale data sets.

To solve the above problems, in this paper, we propose a stable equilibrium point (SEP)-based framework for improving the performance of the above two main EDA tasks. Compared with the existing methods, our method has a nearly-linear time complexity and the SEP-based representative points have strong capability of representing the cluster properties of the original data set. Experimental results show that the proposed method improve the performance
of main EDA tasks for a large margin.

\section{Preliminary}

\subsection{Spectral Clustering Algorithms}
Spectral clustering can often outperform traditional clustering algorithms, such as $k$-means algorithms\cite{von2007tutorial}. Typical spectral clustering algorithms can be divided into three steps: 1) construct a similarity matrix according to the entire data set, 2) embed all data points into $k$-dimensional space using eigenvectors of $k$ bottom nonzero eigenvalues of the graph Laplacian, and 3) perform k-means algorithm to partition the embedded data points into k clusters. Even though these algorithms are rather easy to implement, they are not suitable for handling large-scale data sets since computing eigenvectors of the original graph Laplacian are usually very costly.

\subsection{t-Distributed Stochastic Neighbor Embedding}

t-Distributed Stochastic Neighbor Embedding (t-SNE) \cite{van2008visualizing} maps data points from high-dimensional space to two or three dimension in such a way that similar data points are located in nearby places while dissimilar points are located in distant places. It includes the following main steps: 

1) Convert the euclidean distances into conditional probabilities as follows:

\begin{equation}\label{formula_cond prob}
{P_{j\mid i}} =  {{{\frac{exp(-\frac{{\|x_i-x_j\|}^2}{2{\sigma_i}^2})}{\sum_{k \neq i}exp(-\frac{{\|x_i-x_j\|}^2}{2{\sigma_i}^2})}}}},
\end{equation}

\begin{equation}\label{formula_cond prob2}
{P_{i\mid i}} = 0.
\end{equation}

where $\sigma_i$ denotes the variance of the Gaussian distribution that is centered at $x_i$. Then, the joint probability is defined as follows:

\begin{equation}\label{formula_joint prob high}
{P_{ij}} =  {{{\frac{P_{j\mid i}+P_{i\mid j}}{2{N}}}}}.
\end{equation}

where $N$ is the number of data points in the data set.

2) Assume that  $y_i$ and $y_j$ are two points in the low-dimensional space corresponding to $x_i$ and $x_j$, respectively. The similarity between $y_i$ and $y_j$ is defined as follows:

\begin{equation}\label{formula_joint prob low}
{q_{ij}} =  {{{\frac{(1+{{\|y_i-y_j\|}^2})^{-1}}{\sum_{k \neq l}(1+{{\|y_k-y_l\|}^2})^{-1}}}}},
\end{equation}

\begin{equation}\label{formula_cond low prob2}
{q_{ii}} = 0.
\end{equation}

3) t-SNE uses the sum of Kullback-Leibler divergence over all pairs of data points as the cost function of the dimensionality reduction:

\begin{equation}\label{formula_KL}
{C} = KL({P\parallel Q})= \sum\limits_{i \neq j} {{p_{ij}}} \log{\frac{p_{ij}}{q_{ij}}}.
\end{equation}

For each point, its corresponding point in low-dimensional space is determined by performing gradient descent associated with the above cost function. For a data set with $N$ points, the time complexity of t-SNE is $O(N^2)$.

\section{Methods}

Inspired by the support vector approach, we first map the data from the original space into a high-dimensional space with a nonlinear transformation $\Phi$ to find a sphere with the smallest radius that can enclose all data points in the feature space by solving the following problem:

\begin{equation}
{{\|\Phi(\mathbf{x}_j)-\mathbf{a}\|}}^{2} \leq R^{2}+\xi_j,
\end{equation}

\begin{equation}
\begin{split}
f(\mathbf{x}) &= R^2(\mathbf{x}) = {{\|\Phi(\mathbf{x})-\mathbf{a}\|}}^{2}\\
&= K(\mathbf{x}, \mathbf{x}) - 2\sum_{j} K(\mathbf{x}_j, \mathbf{x}_j)\beta_j + \sum_{i,j} \beta_i \beta_j K(\mathbf{x}_i, \mathbf{x}_j)
\end{split}
\end{equation}

where $K(\mathbf{x}_i, \mathbf{x}_j) = \mathit{e}^{-q \left\| \mathbf{x}_i - \mathbf{x}_j \right\|^2}$ is the Mercer kernel, $\beta_i$ and $\beta_j$ are Lagrangian multipliers.

When the hyper-sphere is mapped back to data space, it forms a set of contours which enclose similar data points. However, these contours cannot be explicitly discovered. But its has been demonstrated that by performing the following gradient descent process associated with $f(\mathbf{x})$, data points that belong to same contour will converge to the same point \cite{lee2005improved}.

\begin{equation}
    \frac{d \mathbf{x}}{dt} = - \nabla f(\mathbf{x}).
\end{equation}

These convergence points are called stable equilibrium points (SEPs). Obviously, SEPs has strong capability of representing a set of similar data points. So, in this paper, we adopt SEPs as the representative points for the data sets.

However, calculating SEPs requires to perform the time consuming gradient descent on every point, which is impractical for large-scale data set. To reduce to computational cost, in this paper, we adopt a novel spectrum-preserving node aggregation-based SEP searching method \cite{song2023accelerate}. Specifically,  a $k$-NN graph $G$ is constructed to capture the global manifold of the data set. Then, the structural correlation of two nodes $p$ and $q$ can be calculated as follows \cite{livne2012lean} :

\begin{equation}\label{eqn:agg}
Sim_{p,q} =\frac{{|(\mathbf{X}_p, \mathbf{X}_q)|}^2}{(\mathbf{X}_p, \mathbf{X}_p)(\mathbf{X}_q, \mathbf{X}_q)},  \mbox{   } (\mathbf{X}_p,\mathbf{X}_q) := \sum_{k=1}^{K}{\left(\mathbf{x}_p^{(k)} \cdot \mathbf{x}_q^{(k)}\right)},
\end{equation}
where $\mathbf{X} := (\mathbf{x}^{(1)}, \dots, \mathbf{x}{^{(K)}})$ is the approximate solution of the bottom $K$ non-trivial eignevectors of the Laplacian matrix ${L}_{G}$ corresponding to the graph $G$. It is obtained by applying Gauss-Seidel relaxation solve ${L}_{G} \mathbf{x}^{(i)}=0$ for $i=1, \dots, K$, starting with $K$ random vectors. 
By aggregating the nodes with high structural correlation, a compressed data set can be formed. We search the SEPs from the compressed data set to reduce the computational cost.

Then, we apply standard spectral clustering on the SEPs to divide them into $k$ clusters. Then, the cluster labels are mapped back from the SEPs to the original data points. Note that here are two levels of mapping. First, we assign the cluster label of each SEP to all of its associated compressed data points. Then, the cluster label of each compressed data point is assigned to all of its associated original data points.The complete algorithm flow has been shown in Algorithm \ref{alg:kaspnew}.

\textbf{Input:} A data set $D$ with $N$ samples $x_1,...,x_N \in {R}^{d}$, number of clusters $k$.\\
\textbf{Output:} Clusters $C_1$,...,$C_k$.\\

\begin{algorithmic}
    \STATE Map the data from the original space into a high-dimensional space to find the minimal enclosing hyper-sphere.
    \STATE Calculate the squared radial distance f(x) of data point x from the hyper-sphere center
    \STATE Search for SEPs by performing gradient descent associated with f(x)
    
    \STATE Perform spectral clustering to divide SEPs into $k$ clusters ; \\
    \STATE Search SEP points from the compressed data set; \\
   
    \STATE Map the cluster-memberships of SEPs back to obtain clustering result of the original data set.\\
    
\end{algorithmic}

This framework can also be applied for achieving fast t-SNE. Recent research [27] shows that the dimensionality reduction process in t-SNE is closely related to the bottom non-trivial eigenvectors of the graph Laplacian matrix corresponding to the data graph. According to the spectral graph theory, the manifold of the data set is encoded in these eigenvectors. This motivates us to propose a SEP-based t-SNE algorithm. Due to the special properties of the SEP, visualizing the representative data set composed of SEPs using the t-SNE will produce similar visualization result with dramatically improved efficiency.

 



\section{Experiment}\label{sect:experiments}
The spectral clustering algorithms and t-SNE are performed using MATLAB running on Laptop. The implementation of the compared methods can be found on their authors' website. 

\subsection{Data sets}

We evaluate the proposed method with three real-world large-scale data sets:\\

1) The \textbf{USPS} data set contains  9,298 16×16 pixel grayscale images of  hand written digits scanned from envelopes of the U.S. Postal Service ;\\

2) The \textbf{MNIST}: The Modified National Institute of Standard and Technology (MNIST) handwritten digits data set. It contains $70,000$ images and each of them is represented by $784$ features; \\

\textbf{Covtype} contains  $581,012$ samples from seven clusters for predicting forest cover type and each sample is represented by $54$ attributes. \\

These three benchmark data set can be downloaded from the UCI machine learning repository \footnote{https://archive.ics.uci.edu/ml/}.

\subsection{Compared Methods}
We compare the proposed method against both the baseline and the state-of-the-art fast spectral clustering methods to demonstrate the effectiveness and efficiency:\\

(1) Standard spectral clustering algorithm \cite{von2007tutorial},\\

(2) Nystr\"om method \cite{fowlkes2004spectral},\\

(3) Landmark-based spectral clustering (LSC) method using random sampling for landmark selection \cite{chen2011large},\\

(4) KASP method using $k$-means for coarse-level clustering \cite{yan2009fast}.\\

For fair comparison, the parameters are set as in \cite{chen2011large} for compared algorithms: the number of sampled points in Nystr\"om method ( or the number of landmarks in LSC, or the number of coarse level cluster centroids in KASP ) is set to 500 as in \cite{wang2021high}.\\

\textbf{Evaluation Metrics.} Clustering accuracy (ACC) is used for evaluating the spectral clustering performance \cite{chen2011large,liu2013large}. The detailed definition can be found in \cite{chen2011parallel}.


\vspace{-0.1in}
\subsection{Experimental Results of Spectral Clsutering}

\begin{table*}[!h]

\begin{center}

\caption{Spectral clustering accuracy (\%)}  
\scalebox{0.9}{
\begin{tabular}{ |c|c|c|c|c|c|c|c|c|c|c|c|c|c|c|c}

 \hline Data Set&Standard SC& Nystr\"om & KASP &LSC&Our Method\\
 
 \hline  PenDigits  &74.36&71.99 &71.56  &74.25&\textbf{76.87}\\
 \hline USPS  &64.31&69.31 &70.62  &66.28&\textbf{75.78}\\

 \hline
\end{tabular}}\label{table:acc result}
\end{center}
\end{table*}

 \begin{table*}[!h]
\begin{center}
\centering
\caption{Runtime (seconds)}  
\scalebox{0.9}{
\begin{tabular}{ |c|c|c|c|c|c|c|c|c|c|c|c|c|c|c|c}

 \hline Data Set&Standard SC&Nystr\"om &KASP  &LSC&Our Method\\
 
 \hline  PenDigits  &0.18&0.19&0.13&0.10&0.07 \\
 \hline USPS  &0.72&0.29&0.16  &0.22 &0.15\\

 \hline
\end{tabular}}\label{table:clustering time pc}
\end{center}
\end{table*}

To show the effectiveness of the proposed method, clustering accuracy results and runtime results are provided in Table 3 and Table 4, respectively. It can be observed that for all the three data sets, the proposed method generates significantly better clustering results than the standard spectral clustering.
This is because our method enables to generate a much better pattern for spectral clustering. Spectral clustering is a graph-based method and it is vert sensitive to noisy nodes in the graph. For example, if only a few noisy nodes form a weak connection between two densely connected clusters, spectral clustering algorithm tends to recognize the two clusters and the noisy nodes as one cluster \cite{bojchevski2017robust}. Compared with standard spectral clustering, the graph in our method is only composed of a small set of SEPs, so that noisy node problem can be fundamentally addressed. Compared with the KASP method, our method is much more efficient. KASP method uses $k$-means centroids as the representative points. In constrast with other method that perform $k$-means on the spectrally-embedded low-dimensional space, the KASP method performs $k$-means in the original high-dimensional data set to generate the representative points.

\subsection{Experimental Results of t-SNE}

Figure 5 shows the visualization and runtime results of the standard t-SNE method and the proposed t-SNE algorithm. It can be seen that we achieve 3 times speedup for USPS data set while displaying clearly cluster structure.

\begin{figure}[htbp]\centering
\includegraphics[scale=0.15]{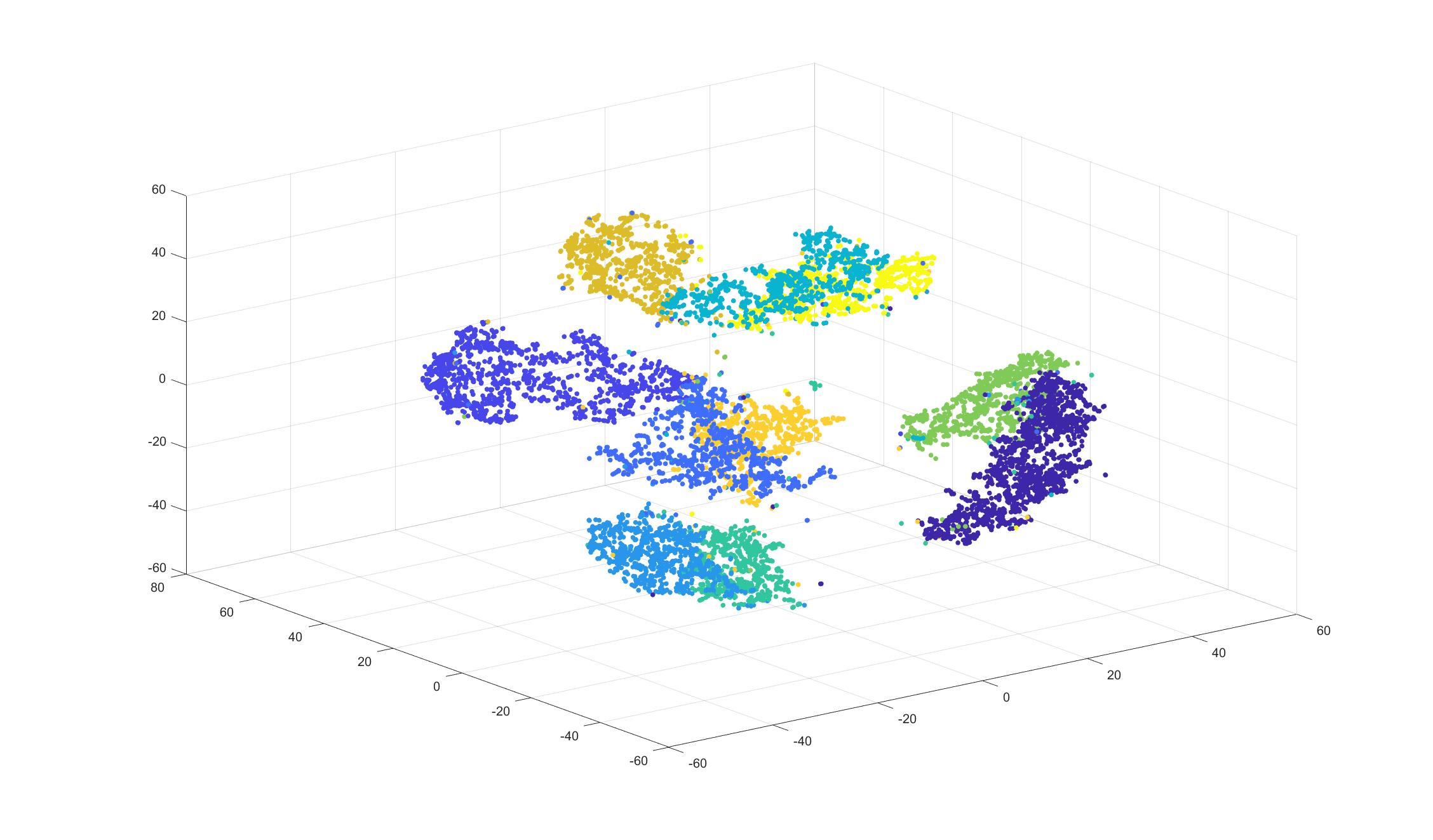}
\caption{Standard t-SNE for the USPS data set (t-SNE time: 77s) .\protect\label{fig:nmi_svc.eps}}
\end{figure}

\begin{figure}[htbp]\centering
\includegraphics[scale=0.15]{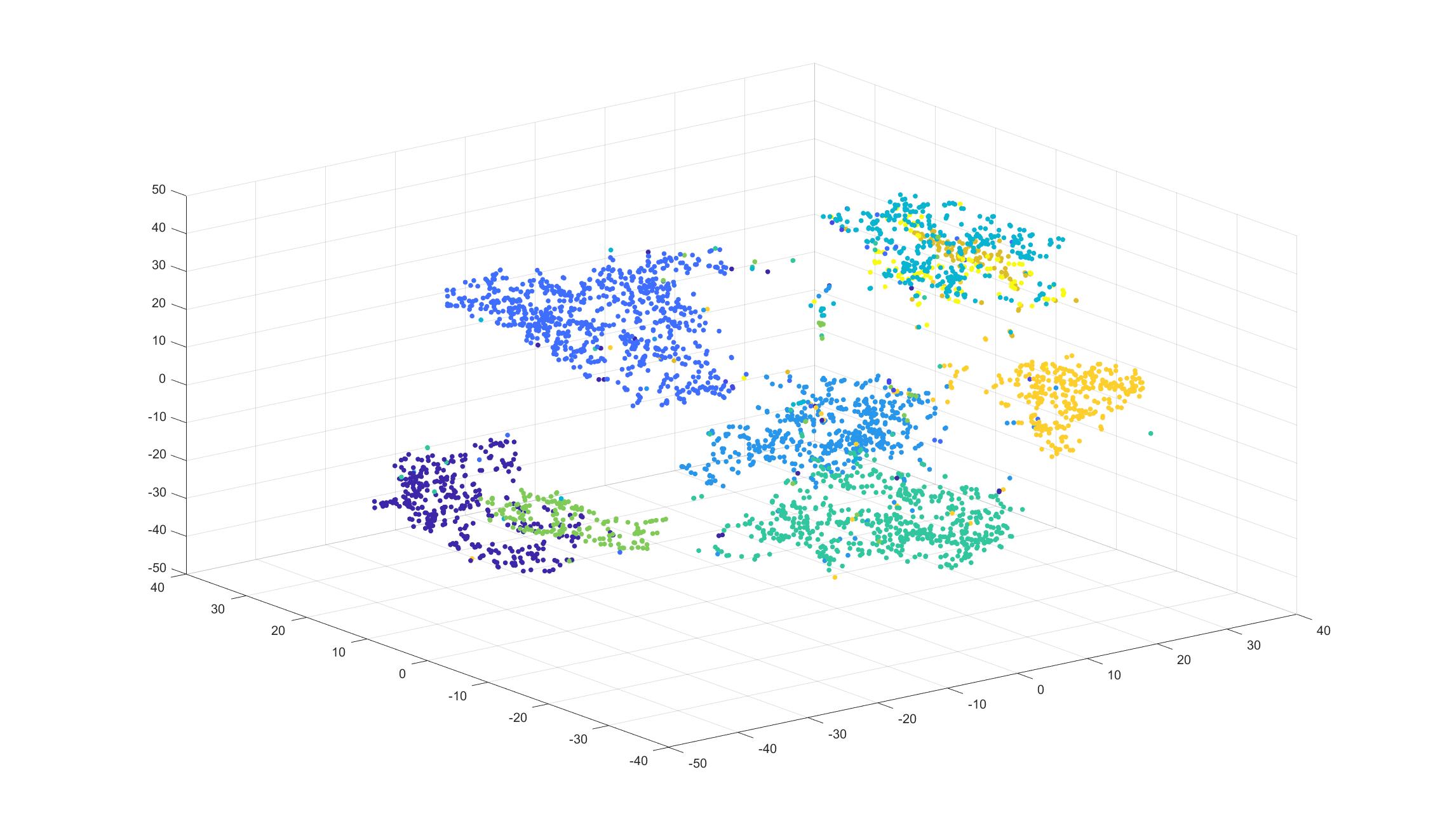}
\caption{SEP-based t-SNE for the USPS data set (t-SNE time: 25s).\protect\label{fig:runtime_svc.eps}}
\end{figure}

\section{Conclusion}\label{sect:conclusions}
We propose a novel stable equilibrium point (SEP) - based algorithmic framework for improving the performance of two main tasks (clustering and data visualization) in exploratory data analysis (EDA). Our method enables to fundamentally address the computational challenge of spectral clustering and t-SNE while improving the solution quality. Experimental results on large scale real-world data sets demonstrate the effectiveness of the proposed methods.

\section{Contributions}\label{sect:conclusions}
Yongyu Wang and Yuxuan Song conceived the idea. Yongyu Wang supervised, led and guided the entire project. Yongyu Wang and Yuxuan Song designed and conducted the experiments and wrote the paper. All authors discussed the results and implications and commented on the manuscript at all stages.


\begin{thebibliography}{8}




\bibitem{yan2009fast}
D.~Yan, L.~Huang, and M.~I. Jordan, ``Fast approximate spectral clustering,''
  in \emph{Proceedings of the 15th ACM SIGKDD international conference on
  Knowledge discovery and data mining}.\hskip 1em plus 0.5em minus 0.4em\relax
  ACM, 2009, pp. 907--916.





\bibitem{fowlkes2004spectral}
C.~Fowlkes, S.~Belongie, F.~Chung, and J.~Malik, ``Spectral grouping using the
  nystrom method,'' \emph{IEEE transactions on pattern analysis and machine
  intelligence}, vol.~26, no.~2, pp. 214--225, 2004.





\bibitem{chen2011large}
X.~Chen and D.~Cai, ``Large scale spectral clustering with landmark-based
  representation.'' in \emph{AAAI}, 2011.

\

\bibitem{von2007tutorial}
U.~Von~Luxburg, ``A tutorial on spectral clustering,'' \emph{Statistics and
  computing}, vol.~17, no.~4, pp. 395--416, 2007.



\bibitem{chen2011parallel}
W.-Y. Chen, Y.~Song, H.~Bai, C.-J. Lin, and E.~Y. Chang, ``Parallel spectral
  clustering in distributed systems,'' \emph{IEEE transactions on pattern
  analysis and machine intelligence}, vol.~33, no.~3, pp. 568--586, 2011.
  









\bibitem{van2008visualizing}
L.~Van~der Maaten and G.~Hinton, ``Visualizing data using t-sne.''
  \emph{Journal of machine learning research}, vol.~9, no.~11, 2008.



\bibitem{lee2005improved}
J.~Lee and D.~Lee, ``An improved cluster labeling method for support vector
  clustering,'' \emph{IEEE Transactions on pattern analysis and machine
  intelligence}, vol.~27, no.~3, pp. 461--464, 2005.

\bibitem{song2023accelerate}
Y.~Song and Y.~Wang, ``Accelerate support vector clustering via
  spectrum-preserving data compression?'' \emph{arXiv preprint
  arXiv:2304.09868}, 2023.




\bibitem{wang2021high}
Y.~Wang, ``High performance spectral methods for graph-based machine
  learning,'' Ph.D. dissertation, Michigan Technological University, 2021.

\bibitem{livne2012lean}
O.~E. Livne and A.~Brandt, ``Lean algebraic multigrid (lamg): Fast graph
  laplacian linear solver,'' \emph{SIAM Journal on Scientific Computing},
  vol.~34, no.~4, pp. B499--B522, 2012.






\end{thebibliography}
\end{document}